\documentclass{article}
\usepackage{ijcai25}

\usepackage{times}
\usepackage{soul}
\usepackage{url}
\usepackage[hidelinks]{hyperref}
\usepackage[utf8]{inputenc}
\usepackage[small]{caption}
\usepackage{graphicx}
\usepackage{amsmath}
\usepackage{amsthm}
\usepackage{booktabs}
\usepackage{algorithm}
\usepackage{algorithmic}
\usepackage[switch]{lineno}

\usepackage{multirow}
\usepackage{arydshln}
\usepackage{epsfig}
\usepackage{pifont}
\usepackage{rotating}

\newcommand{\ie}{\textit{i}.\textit{e}.}
\newcommand{\eg}{\textit{e}.\textit{g}.}
\newcommand{\etal}{\textit{et al}.}

\usepackage{amsfonts}

\title{Stochasticity-aware No-Reference Point Cloud Quality Assessment}

\author{
Songlin Fan$^{1,3}$ \and
Wei Gao$^{1}$\footnote{Corresponding author.} \and
Zhineng Chen$^{2}$ \and
Ge Li$^{1}$ \and
Guoqing Liu$^{4}$ \And
Qicheng Wang$^{4}$ \\
\affiliations
$^1$Guangdong Provincial Key Laboratory of Ultra High Definition Immersive Media Technology, Shenzhen Graduate School, Peking University\\
$^2$School of Computer Science, Fudan University \\
$^3$China Mobile Shanghai ICT Co., Ltd\\
$^4$Youiia Innov Tech Co., Ltd
\emails
\{slfan, gaowei262, geli\}@pku.edu.cn, zhinchen@fudan.edu.cn, \{guoqing, wangqicheng\}@minieye.cc
}

\begin{document}

\maketitle

\begin{abstract}
The evolution of point cloud processing algorithms necessitates an accurate assessment for their quality. Previous works consistently regard point cloud quality assessment (PCQA) as a MOS regression problem and devise a deterministic mapping, ignoring the stochasticity in generating MOS from subjective tests.  This work presents the first probabilistic architecture for no-reference PCQA, motivated by the labeling process of existing datasets. The proposed method can model the quality judging stochasticity of subjects through a tailored conditional variational autoencoder (CVAE) and produces multiple intermediate quality ratings. These intermediate ratings simulate the judgments from different subjects and are then integrated into an accurate quality prediction, mimicking the generation process of a ground truth MOS. Specifically, our method incorporates a Prior Module, a Posterior Module, and a Quality Rating Generator, where the former two modules are introduced to model the judging stochasticity in subjective tests, while the latter is developed to generate diverse quality ratings. Extensive experiments indicate that our approach outperforms previous cutting-edge methods by a large margin and exhibits gratifying cross-dataset robustness. Codes are available at \url{https://git.openi.org.cn/OpenPointCloud/nrpcqa}.
\end{abstract}

\section{Introduction}
Point clouds are essential representations of 3D scenes and have found wide applications in various fields~\cite{qi2017pointnet,fan2024point}. Due to the unaffordable complexity and size of raw point clouds, compression techniques are commonly used to reduce relevant processing and storage costs. However, compression can lead to visual quality degradation. Therefore, accurate measurement of point cloud quality is crucial to assess the fidelity of processed data and serves as a foundation for benchmarking and improving point cloud processing algorithms. Similar to the definitions for image quality assessment, point cloud quality assessment (PCQA) can be categorized into three factions: full-reference PCQA (FR-PCQA), reduced-reference PCQA (RR-PCQA), and no-reference PCQA (NR-PCQA), depending on the availability of reference data. As pristine data are often transparent in typical client applications, NR-PCQA has received significant attention and exploration efforts.

\begin{figure}[t]
\centering
\includegraphics[width=1\linewidth]{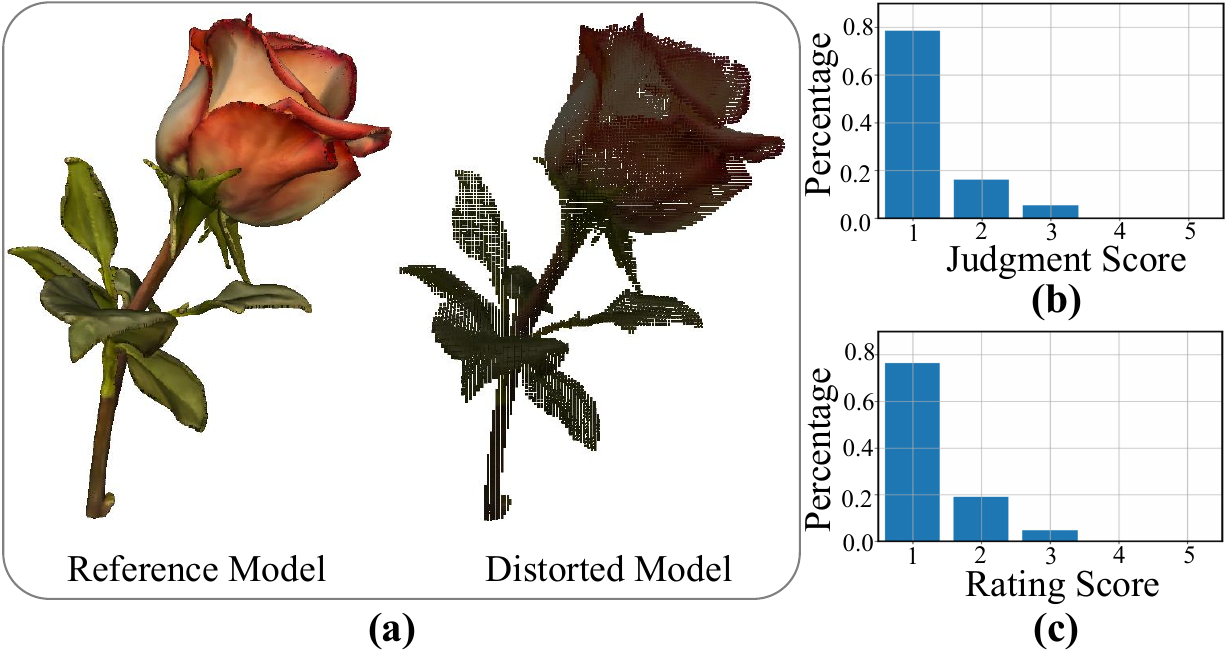}
\caption{Stochasticity in standard dataset construction and predictions of our method. (a) Paired stimuli. (b) Distribution of 37 quality judgments about (a) in a recent subjective test. (c) Distribution of 37 intermediate quality ratings predicted from our model.}
\label{fig1}
\end{figure}

Conventional NR-PCQA methods~\cite{wang2023applying,zhang2022nrqa,liu2021pqa,zhang2022mm,chetouani2021deep,liu2023point} consistently adhere to a standardized framework that devises a deterministic fitting function or neural network mapping ${M}=f(I;\omega)$, where $\omega$ denotes the model parameters/weights, $I$ represents the input point cloud or its derivatives (\eg, patches or projections), and $M$ corresponds to the mean opinion score (MOS). MOSs, the most reliable visual quality description for point clouds, are usually obtained via subjective tests~\cite{fan2023screen,liu2022perceptual,liu2021reduced,yang2020predicting}. A subjective test involves recruiting several subjects to conduct laborious subjective experiments and gather human judgments for the viewed point cloud stimuli. Eventually, the discrepant judgment scores of a sample from different subjects are synthesized or directly averaged into the final MOS. We argue that though distortions in point clouds are deterministic, existing NR-PCQA solutions deviate from the generation of ground truth MOSs in subjective tests and fail to model the stochasticity in the human visual system (HVS) when giving point cloud quality judgements.

Results of subjective tests~\cite{fan2023screen,liu2023point,wu2021subjective,javaheri2020point} reveal that the HVS exhibits inherent volatility and stochasticity. That is, the quality judgments of different subjects may vary when evaluating the same point cloud stimuli due to the subjective nature of the HVS, which is influenced by complex factors such as individual visual sensitivity, cognitive biases, and perceptual preferences. Even a single subject may have distinct judgments for the same stimuli due to temporal fluctuations induced by prior context and mood. Figure~\ref{fig1} illustrates the stochasticity of a standard subjective test~\cite{fan2023screen}, showing that the actual quality judgments of a point cloud given by different subjects exhibit a specific distribution rather than a deterministic opinion. Herefore, existing overconfident methods that rely on deterministic mappings are susceptible to judgment biases. Furthermore, fitting a potentially biased MOS is one of the most critical factors that weaken the robustness of existing methods. Inspired by practical subjective tests, this work presents the first attempt at judgment distribution prediction and models the stochasticity in ground truth MOSs to produce multiple quality ratings for each point cloud. Multiple ratings are then integrated into a comprehensive quality prediction.

In specific, we propose a novel conditional variational autoencoder (CVAE) architecture for NR-PCQA, which leverages a latent variable conditioned on the projections of point clouds to accommodate the judgment variations of different subjects. Since the raw quality judgments in subjective tests of popular datasets are unavailable, which is necessary for training a CVAE to avoid posterior collapse, we utilize an effective KL annealing strategy~\cite{sonderby2016ladder} to achieve diverse quality outputs for a point cloud. The proposed CVAE architecture comprises a Prior Module ($\text{PM}_{\text{prior}}$), a Posterior Module ($\text{PM}_{\text{post}}$), and a Quality Rating Generator (QRG). The $\text{PM}_{\text{prior}}$ and $\text{PM}_{\text{post}}$ are developed to predict the prior and posterior distributions of the latent variable, from which stochastic features can be sampled. Each sampled stochastic feature expresses the factors that may result in a possible judgment variant in subjective tests. Given the sampled stochastic feature, we further introduce the QRG to excavate the uncertain information in the stochastic feature and deterministic distortion information in point cloud projections and produce a stochastic quality rating. As the prior and posterior distribution mismatch of CVAEs can merely guarantee suboptimal performance~\cite{sohn2015learning}, we utilize the Gaussian Stochastic Neural Network (GSNN)~\cite{sohn2015learning} to mitigate the disparity between training and testing. In the testing phase, we can sample the stochastic feature multiple times to predict diverse quality ratings of a point cloud. As shown in Figure~\ref{fig1}, the diverse rating outputs of our method well capture the actual distribution of judgments derived from subjective tests, which can finally be averaged into an accurate quality prediction. Experimental results demonstrate that our method significantly outperforms all compared counterparts, including FR-PCQA and NR-PCQA methods.

We highlight the contributions of this work as follows:
\begin{itemize}
\item We are the first to probe stochasticity in dataset labeling for PCQA and propose a probabilistic perspective mimicking subjective tests and patterning the quality judgment distribution rather than a potentially biased MOS.
\item We are the first to utilize a tailored CVAE in the quality assessment field, which can model both the uncertain factors in subjective tests and deterministic distortions in point clouds for accurate quality prediction.
\item We verify the effectiveness of our method by achieving a new state-of-the-art performance and gratifying cross-dataset robustness, surpassing the existing best method by a significant margin across all datasets.
\end{itemize}

\begin{figure*}[t]
\centering
\includegraphics[width=\linewidth]{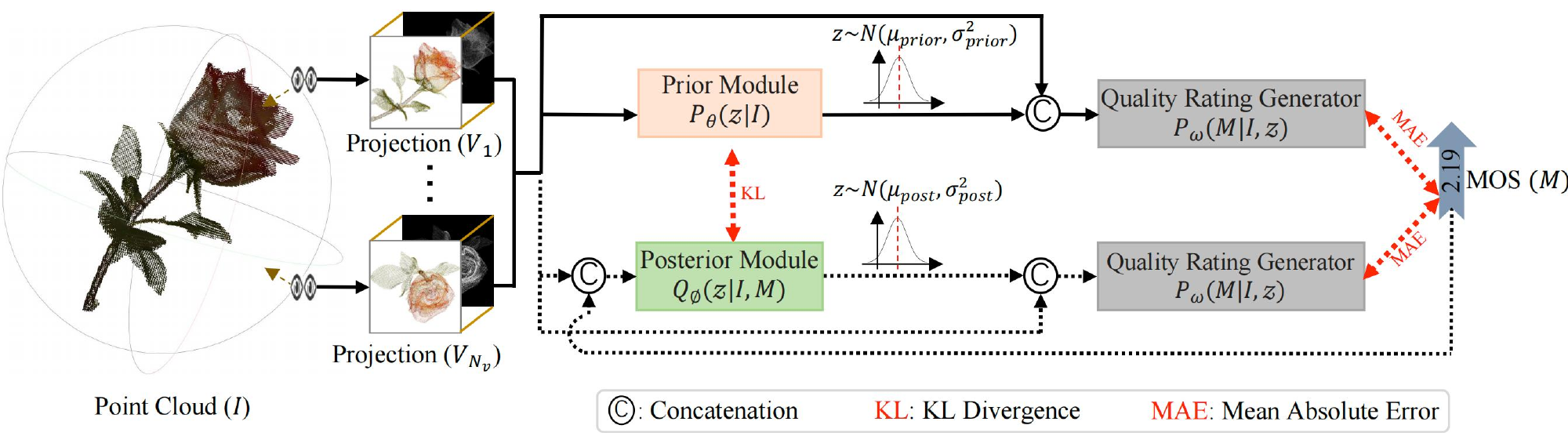}
\caption{Overall architecture of the proposed method.  Our method comprises a Prior Module ($\text{PM}_{\text{prior}}$), a Posterior Module ($\text{PM}_{\text{post}}$), and a Quality Rating Generator (QRG). Note that the two QRGs share network parameters. Solid lines indicate functionality in both the training and testing phases, while dashed lines indicate only working in the training stage. }
\label{fig2}
\end{figure*}

\section{Related Work}
\subsection{Point Cloud Quality Assessment}
The surging popularity of point clouds has sparked significant interest in PCQA. The development of PCQA has drawn inspiration from the advancements in image quality assessment, leading to categorizing methods into FR-PCQA, RR-PCQA, and NR-PCQA concerning their reliance on pristine point clouds.  Early interest in FR-PCQA emerges from MPEG, with a primary focus on evaluating and optimizing point cloud compression algorithms. For example, point-to-point error~\cite{mekuria2016evaluation} and point-to-plane error~\cite{tian2017geometric} are proposed to measure the geometry quality of point clouds and lay a solid foundation for subsequent investigations~\cite{alexiou2018point,javaheri2020generalized,meynet2019pc,yang2020predicting}. Inspired by the widely-used metrics in images (\ie, PSNR and SSIM), Torlig~\etal~\cite{torlig2018novel} and Alexiou~\etal~\cite{alexiou2020towards} design the point cloud-based versions, respectively, namely PSNR-yuv and PointSSIM. Yang~\etal~\cite{yang2020inferring} infer the perceptual quality of point clouds using graph signal gradient and construct local graph representations. PCQM~\cite{meynet2020pcqm} considers both geometry and color distortions and utilizes a weighted combination to calculate a comprehensive quality description of point clouds. Recently, a novel distortion quantification method is devised by Yang~\etal~\cite{yang2021point} to model point cloud quality via multi-scale potential energy discrepancy. Apart from the aforementioned FR-PCQA works, some notable RR-PCQA methods~\cite{liu2021reduced,liu2022reduced} are also proposed and enrich the research on PCQA.

As considerable applications cannot observe reference point clouds, NR-PCQA~\cite{wang2023applying} has attracted lots of interest. PQA-Net~\cite{liu2021pqa} is the first NR-PCQA method that divides the PCQA problem into two meaningful sub-tasks, including distortion classification and quality prediction. These two sub-tasks cooperate to obtain gratifying results. Chetouani~\etal~\cite{chetouani2021deep} adopt conventional deep learning routes on patch-level distortion characterizations to regress the quality scores of input samples. 3D-NSS~\cite{zhang2022no} utilizes support vector regression (SVR) to regress the quality-aware feature and obtain the visual quality score, which can address both point cloud and mesh quality assessment problems. Fan~\etal~\cite{fan2022no} first capture three videos by rotation operations and then conduct the PCQA task based on video features. Liu~\etal~\cite{liu2023point} present an effective quality metric, termed ResSCNN, which can accurately estimate MOSs of point clouds. Besides, they also construct the largest-scale dataset. To promote the robustness of existing NR-PCQA models, Yang~\etal~\cite{yang2022no} leverage existing abundant subjective scores of natural images and transfer relevant knowledge to help point cloud quality recognition. All these discussed NR-PCQA methods treat their tasks as a MOS fitting problem and fail to take the stochasticity in the MOS generation process (subjective tests) into account. 

\subsection{Generative Model}
Variational autoencoders (VAE)~\cite{kingma2013auto} and conditional variational autoencoders (CVAE)~\cite{sohn2015learning} are types of generative models that combine the principles of autoencoders and variational inference. VAEs extend traditional autoencoders by incorporating probabilistic modeling and typically map the input to a standard Gaussian distribution. The training process of VAEs involves maximizing a lower bound on data log-likelihood while minimizing the Kullback-Leibler (KL) divergence between the prior and posterior distributions of latent features. CVAEs are a further extension of VAEs, which modulate the Gaussian distribution of latent features by integrating conditional information. Due to their outstanding performance, the frameworks of VAEs and CVAEs have been widely used in various computer vision tasks. Liu~\etal~\cite{liu2021variational} design a novel RefVAE for reference-based image super-resolution, while Esser~\etal~\cite{esser2018variational} apply the VAE to the image generation task. Recently, we also observed that some innovative works~\cite{baumgartner2019phiseg,zhang2021uncertainty}  introduce CVAEs into segmentation tasks and achieve visibly improved performance. 

In the quality assessment field, generative adversarial networks (GAN) as another type of generative models have been exploited by some works~\cite{lin2018hallucinated,ma2020active,zhu2021recycling,ren2018ran4iqa,yang2020ttl} to assess image quality. Moreover, most of these works employ the framework of GANs to generate pseudo-reference images or for transfer learning purposes. To the best of our knowledge, no work explores the utilization of VAEs or CVAEs in the context of quality assessment.

\section{Proposed Method}

As shown in Figure~\ref{fig1}, our method aims to learn the conditional probability distribution of quality judgments in subjective tests instead of a solitary and potentially biased judgment. With a training dataset $\mathcal{D}=\{I_i,M_i\}_{i=1}^{N_s}$ comprising $N_s$ point clouds, we endeavor to estimate the distribution $P_\omega(M|I,z)$, where $I$, $M$, and $z$ represent the point cloud sample, corresponding MOS, and a low-dimensional latent variable, respectively. The expression $P_\omega(M|I,z)$ indicates that the quality judgment from a subject depends on both the deterministic distortions in point cloud $I$ and the uncertain factors $z$ influenced by the HVS. Our proposed framework, illustrated in Figure~\ref{fig2}, has three modules: a Prior Module ($\text{PM}_{\text{prior}}$), a Posterior Module ($\text{PM}_{\text{post}}$), and a Quality Rating Generator (QRG). During training, the $\text{PM}_{\text{post}}$ approximates the posterior distribution of the latent variable $Q_\phi(z|I,M)$ and aids the learning of  $\text{PM}_{\text{prior}}$ regarding the prior distribution $P_\theta(z|I)$. In the testing phase, stochastic features are sampled from the prior distribution, which, along with point cloud projections, are input to the QRG for predicting stochastic quality ratings.

As investigated by Liu~\etal~\cite{liu2022perceptual}, most previous subjective tests for annotating PCQA datasets adopt an interactive 2D monitor as their display tool. To exactly agree with previous subjective tests, our model projects each point cloud into $N_v$ four-channel RGB-D projections in the input stage, similar to the previous work~\cite{yang2020predicting}.  We define the $N_v$ four-channel projections from a point cloud as $\{V_i\}_{i=1}^{N_v}$. When computing projections, unlike previous methods~\cite{yang2020predicting,liu2021pqa,fan2022no,zhang2022mm} that employ fixed viewpoints, our model allows for the random selection of viewpoints to imitate the interactive viewing pattern in subjective tests, where any view can be used to judge the quality of a point cloud.  This random viewpoint selection strategy not only provides stochastic priors in the stochasticity modeling process, but also enhances the robustness of neural networks, since fixing viewpoint selection can be treated as a special case of our random viewpoint selection strategy.

\subsection{Problem Formulation}
CVAEs often incorporate a conditioning variable, an output variable, and a latent variable. In the PCQA task, the conditioning and output variables correspond to the input point cloud $I$ and the predicted quality rating $M$, respectively. The prior distribution of the latent variable $z$ follows a modulated Gaussian distribution $P_\theta(z|I)$, with its parameters conditioned on the input point cloud $I$. The quality rating $M$ is derived from $P_\omega(M|I,z)$, leading to the posterior distribution of $z$ expressed as $Q_\phi(z|I,M)$. The primary objective of CVAEs is to approximate the posterior distribution of the latent variable, given the conditioning information. This is achieved by maximizing the evidence lower bound (ELBO). Specifically, the loss function of standard CVAEs consists of a reconstruction loss and a regularizer, which can be represented as:
\begin{equation}
\label{eq1}
\begin{aligned}
\mathcal{L}_{CVAE}= \mathbb{E}_{z\sim Q_\phi(z|I,M)}[-logP_\omega(M|I,z)]\\+D_{KL}(Q_\phi(z|I,M)||P_\theta(z|I)),
\end{aligned}
\end{equation}
where the first term denotes the reconstruction loss, which quantifies the negative conditional log-likelihood of the output variable $M$, given the conditioning variable $I$ and the latent variable $z$ drawn from $Q_\phi(z|I,M)$. The second term represents the KL divergence of $Q_\phi(z|I,M)$ and $P_\theta(z|I)$, acting as a regularizer to minimize the distribution discrepancy between the prior and posterior distributions.

However, training CVAEs to generate diverse quality ratings for a point cloud requires annotating training samples with multiple versions of quality judgments, as manifested by the varying judgments obtained from different subjects in subjective tests. Unfortunately, existing popular datasets~\cite{yang2020predicting,liu2022perceptual,liu2021reduced} only provide a single MOS for each sample without disclosing the raw quality judgments obtained in subjective tests. Consequently, directly employing the conventional CVAE architecture with these datasets may lead to potential posterior collapse, rendering the latent variable invalid and making the training process unstable. To address the issue of posterior collapse, we adopt an effective KL annealing strategy~\cite{sonderby2016ladder} during training, gradually increasing the weight of the regularizer in Equation~(\ref{eq1}). This strategy can be described as:
\begin{equation}
\label{eq2}
\begin{aligned}
\mathcal{L}_{CVAE}^{\lambda} = \mathbb{E}_{z\sim Q_\phi(z|I,M)}[-logP_\omega(M|I,z)]\\+\lambda*D_{KL}(Q_\phi(z|I,M)||P_\theta(z|I)),
\end{aligned}
\end{equation}
where $\lambda$ equals the ratio of the current training epoch to the total training epoch. As depicted in Figure~\ref{fig1}, the model with our KL annealing strategy enables diverse quality rating outputs for a point cloud, despite being trained on samples with merely a single annotation.

\begin{figure}[t]
\centering
\includegraphics[width=0.95\linewidth]{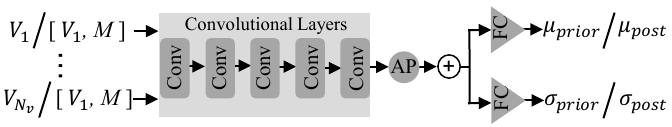}
\caption{Network structure of the Prior Module and the Posterior Module. “[$\cdot$]” denotes the concatenated input of the Posterior Module. “Conv” represents the convolutional layer. “$\oplus$” is element-wise summation.
“AP” means the global average pooling layer. “FC” stands for the fully connected layer. }
\label{fig3}
\end{figure}

\subsection{Prior/Posterior Module}
With the adapted objective of CAVEs for PCQA in Equation~(\ref{eq2}), we utilize two modules, the Prior Module ($\text{PM}_{\text{prior}}$) and the Posterior Module ($\text{PM}_{\text{post}}$), to model the prior $P_\theta(z|I)$ and posterior $Q_\phi(z|I,M)$ distributions of the latent variable $z$. The two modules share an identical network structure but possess independent parameters, $\theta$ and $\phi$, separately. The $\text{PM}_{\text{prior}}$ is designed to map the four-channel projections of a point cloud $\{V_i\}_{i=1}^{N_v}$ to a low-dimensional prior statistic. To this end, we first employ five convolutional layers, as illustrated in Figure~\ref{fig3}, to extract the individual latent feature from each projection. Subsequently, the latent feature is spatially averaged into a vector representation, and the vector representations of different projections are aggregated via element-wise summation. We map the aggregated vector representation to the Gaussian prior statistic $P_\theta(z|I)$ using two fully connected layers, which is achieved by predicting the mean vector $\mu_{prior}\in\mathbb{R}^{K_1}$ and standard deviation vector $\sigma_{prior}\in\mathbb{R}^{K_1}$ of a Gaussian distribution. Differing from the $\text{PM}_{\text{prior}}$, the $\text{PM}_{\text{post}}$ takes the five-channel concatenation of each point cloud projection and the spatially-expanded MOS as input, inferring the $\mu_{post}, \sigma_{post}\in\mathbb{R}^{K_1}$ of the Gaussian posterior distribution $Q_\phi(z|I,M)$.

During training, the adapted CVAE maximizes the conditional log-likelihood of quality ratings while alleviating the distribution mismatch measured by $D_{KL}(Q_\phi(z|I,M)||P_\theta(z|I))$. As illustrated in Figure~\ref{fig2}, we use the reparameterization trick to sample the stochastic feature $F_s\in\mathbb{R}^{K_1\times H\times W}$ from the posterior statistic $Q_\phi(z|I,M)$, where $H$ and $W$ are the height and width of point cloud projections. Specifically, we generate a $K_1$-dimensional independent random variable $\epsilon$, in which each element is drawn from a standard Gaussian distribution. The reparameterized feature vector $z^s\in\mathbb{R}^{K_1}$ is then computed as $z^s = \sigma_{post}*\epsilon+\mu_{post}$. To obtain the stochastic feature $F_s$, we expand the spatial dimensions of $z^s$ to match the point cloud projections. Ultimately, the stochastic feature $F_s$ is combined with the projection set $\{V_i\}_{i=1}^{N_v}$ to produce the quality rating prediction. In the testing phase, the stochastic feature is sampled from the prior statistic $P_\theta(z|I)$ in the same way to predict a stochastic quality rating.

\subsection{Quality Rating Generator}
\begin{figure}[t]
\centering
\includegraphics[width=0.95\linewidth]{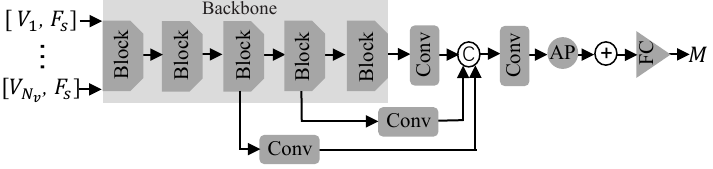}
\caption{Illustration of our Quality Rating Generator. “Block” represents the convolutional block of the backbone. }
\label{fig4}
\end{figure}

The stochastic feature $F_s$ encodes uncertain variability in subjective tests, while the point cloud projections $\{V_i\}_{i=1}^{N_v}$ incarnate the deterministic point cloud distortions. We further develop an effective Quality Rating Generator (QRG) to mimic a subject that associates the uncertain and deterministic information to give quality judgments in subjective tests. Our QRG (see Figure~\ref{fig4}) is built on the ResNet-50~\cite{he2016deep} backbone, whose input is the concatenation of each point cloud projection and the stochastic feature. We use the last three levels of features to compute point cloud quality. Specifically, we first introduce three convolutional layers to reduce the channel size of these three high-level features to $K_2$ and then fuse them into a multi-scale feature by the concatenation operation followed by convolution. The effectiveness of multi-scale representations has been validated in various computer vision tasks~\cite{gao2019res2net,gao2023thorough}. The multi-scale feature is spatially averaged into a vector representation, and we integrate the vector representations from different projections into the quality-ware feature using element-wise summation. The final stochastic quality rating can be predicted from the quality-ware feature via the fully connected layer. During testing, we can sample the stochastic feature multiple times from the prior statistic and produce diverse stochastic ratings for a point cloud. Similar to the calculation of ground truth MOSs, we treat the averaged score of multiple ratings as the final quality prediction for a point cloud.  

\begin{table*}[th]\small
\renewcommand{\tabcolsep}{1.4mm}
\centering
\begin{tabular}{lc|cccccccccccc}
\toprule
  &\multirow{2}{*}{Method}  & \multicolumn{4}{c}{SJTU-PCQA~\cite{yang2020predicting}} & \multicolumn{4}{c}{WPC~\cite{liu2022perceptual}} & \multicolumn{4}{c}{WPC2.0~\cite{liu2021reduced}}\\ 
\cmidrule(l){3-6} \cmidrule(l){7-10} \cmidrule(l){11-14}
        & & SRCC$\uparrow$       & PLCC$\uparrow$      & KRCC$\uparrow$     & RMSE$\downarrow$     & SRCC$\uparrow$      & PLCC$\uparrow$      & KRCC$\uparrow$       & RMSE$\downarrow$  & SRCC$\uparrow$      & PLCC$\uparrow$      & KRCC$\uparrow$       & RMSE$\downarrow$ \\ \midrule\midrule
\multirow{8}{*}{\begin{sideways}{FR-PCQA}\end{sideways}}
 & MSE-p2po & 0.7294 & 0.8123 & 0.5617 & 0.1361 & 0.4558 & 0.4852 & 0.3182 & 0.1989 & 0.4315 & 0.4626 & 0.3082 & 0.1916  \\
 & HD-p2po & 0.7157 & 0.7753 & 0.5447 & 0.1448 &0.2786 &0.3972&0.1943 &0.2090 &0.3587 & 0.4561 & 0.2641 & 0.1890   \\
  & MSE-p2pl & 0.6277 & 0.5940 & 0.4825 & 0.2282 & 0.3281 & 0.2695 &0.2249 & 0.2282 & 0.4136 & 0.4104 & 0.2965 & 0.2104     \\
 & HD-p2pl  & 0.6441   & 0.6874    & 0.4565 & 0.2126 & 0.2827 & 0.2753 &0.1696 &0.2199  & 0.4074 & 0.4402 & 0.3174 & 0.1952  \\
 & PSNR-yuv  & 0.7950 & 0.8170 & 0.6196 & 0.1315 & 0.4493 & 0.5304 & 0.3198 & 0.1931 & 0.3732 & 0.3557 & 0.2277 & 0.2015\\
 & PCQM     & {0.8644}   & {0.8853}    & {0.7086}     & {0.1086}     &{0.7434}    &{0.7499}   &{0.5601}   & 0.1516    &0.6825 & 0.6923 & 0.4929 & 0.1563            \\
 & GraphSIM  & {0.8783}    &{0.8449}    & {0.6947}   &{0.1032}  & 0.5831    & 0.6163    & 0.4194   & 0.1719 & 0.7405 & 0.7512 &{0.5533} &{0.1499}\\
& PointSSIM    & 0.6867  & 0.7136  & 0.4964 & 0.1700  & 0.4542    & 0.4667    & 0.3278   & 0.2027  & 0.4810 & 0.4705 & 0.2978 & 0.1939 \\ \hdashline
\multirow{7}{*}{\begin{sideways}{NR-PCQA}\end{sideways}} 
&BRISQUE  & 0.3975    & 0.4214  & 0.2966 & 0.2094  & 0.2614    & 0.3155  & 0.2088 & 0.2117  & 0.0820  & 0.3353	& 0.0487 & 0.2167
\\
&NIQE  &0.1379 &0.2420 &0.1009 &0.2262  &0.1136 & 0.2225 &0.0953 &0.2314 & 0.1865 & 0.2925 & 0.1335 & 0.2251
 \\
&IL-NIQE  & 0.0837 & 0.1603 & 0.0594 &0.2338 &0.0913 &0.1422 & 0.0853 &0.2401 & 0.0911 & 0.1233 & 0.0714 & 0.2400\\
&ResSCNN & 0.8600 & 0.8100 & - & - & - & - & - & - &{0.7500} &{0.7200} &-&- \\
&PQA-Net & 0.8372   & 0.8586    & 0.6304 & {0.1072}  & {0.7026}    & {0.7122}    & {0.4939}   &{0.1508} & 0.6191 & 0.6426 & 0.4606 & 0.1698   \\
&3D-NSS  & 0.7144 & 0.7382  & 0.5174 & 0.1769    & 0.6479    & 0.6514    & 0.4417   & 0.1657 & 0.5077 & 0.5699 & 0.3638 & 0.1772 \\ 
&MM-PCQA  &{0.9103} &{0.9226} &{0.7838} &{0.0772} &{0.8414}  &{0.8556}  &{0.6513} &{0.1235} &{0.8023}  &{0.8024} &{0.6202}  &{0.1343} \\
&\textbf{Ours} &\textbf{0.9474}&\textbf{0.9636}&\textbf{0.8192}&\textbf{0.0628} &\textbf{0.8744} &\textbf{0.8766} &\textbf{0.6901} &\textbf{0.1089} &\textbf{0.8742} &\textbf{0.8880} &\textbf{0.6922} &\textbf{0.0979}\\

\bottomrule
\end{tabular}
\caption{Benchmarking results of state-of-the-art methods on the SJTU-PCQA, WPC, and WPC2.0 datasets. $``\uparrow"/``\downarrow"$ indicates that larger/smaller is better. The best method is highlighted in \textbf{bold}.
}
\label{tab1}
\end{table*}

\begin{table}[t]\small
\renewcommand{\tabcolsep}{2.8mm}
\centering
\begin{tabular}{l|cccc}
\toprule
\multirow{2}{*}{Method} &\multicolumn{2}{c}{WPC$\to$ SJTU} &\multicolumn{2}{c}{WPC$\to$ WPC2.0}\\
\cmidrule(l){2-3} \cmidrule(l){4-5}
&SRCC$\uparrow$ &PLCC$\uparrow$ &SRCC$\uparrow$ &PLCC$\uparrow$ \\
\midrule\midrule
PQA-Net&0.5411&0.6102&0.6006&0.6377 \\
3D-NSS&0.1817&0.2344&0.4933&0.5613 \\
MM-PCQA&0.7693&0.7779&0.7607&0.7753 \\
\textbf{Ours}&\textbf{0.9133}&\textbf{0.9404}&\textbf{0.8300}&\textbf{0.8511} \\
\bottomrule
\end{tabular}
\caption{Result comparisons of cross-dataset generalization.}  
\label{tab2}
\end{table}

\subsection{Objective Function}
Conventional CVAEs often sample the posterior statistic $Q_\phi(z|I,M)$ for reconstruction during training while relying on the prior statistic $P_\theta(z|I)$ during testing. However, it has been proved~\cite{sohn2015learning} that the distribution mismatch between $Q_\phi(z|I,M)$ and $P_\theta(z|I)$ may lead to suboptimal performance in the testing phase. Although increasing the weight $\lambda$ of the regularizer in Equation~(\ref{eq2}) can mitigate the issue of distribution mismatch, this solution cannot bring overall performance improvement due to the relatively insufficient emphasis on result reconstruction. Inspired by GSNN~\cite{sohn2015learning}, we introduce the “testing branch” (see Figure~\ref{fig2}) to reflect the testing circumstances during training. Concretely, we additionally sample the stochastic feature from $P_\theta(z|I)$ for another reconstruction process in the training stage, which can be formulated as:
\begin{equation}
\label{eq3}
\begin{aligned}
\mathcal{L}_{GSNN} = \mathbb{E}_{z\sim P_\theta(z|I)}[-logP_\omega(M|I,z)].
\end{aligned}
\end{equation}
In this approach, we can force the $\text{PM}_{\text{prior}}$ to learn an informative prior statistic and alleviate the distribution mismatch of the latent variable during the training and testing stages.

Consequently, the overall objective function of our PCQA-tailored CVAE architecture during training is composed of two components, including the adapted CAVE loss $\mathcal{L}_{CVAE}^{\lambda}$ and the GSNN loss $\mathcal{L}_{GSNN}$, which can be expressed as: 
\begin{equation}
\label{eq4}
\begin{aligned}
\mathcal{L}_{overall} = \alpha*\mathcal{L}_{CVAE}^{\lambda} + (1-\alpha)*\mathcal{L}_{GSNN},
\end{aligned}
\end{equation}
where $\alpha$ is a weighting factor for balancing the two loss terms. Furthermore, we use the Mean Absolute Error (MAE) to measure the reconstruction loss between the predicted stochastic quality ratings and ground truth MOSs.

\section{Experiments}
\subsection{Implementation Details}
We implement our model on two NVIDIA RTX 3090 Ti GPUs with the PyTorch toolbox and initialize the ResNet-50~\cite{he2016deep} backbone in the QRG with parameters pre-trained on ImageNet, while other neural network parameters are randomly initialized. We use the Adam optimizer with an initial learning rate of 2.5e-5 and betas set to [0.5, 0.999]. Our model is trained for a total of 200 epochs, and the learning rate is reduced by a factor of 0.5 when the training process reaches the halfway mark. We set the training batch size to 8 while convincing ablations demonstrate that the weighting term $\alpha=0.4$ to emphasize the disparity reduction between the training and testing stages can obtain the best performance. The spatial resolution of point cloud projections is $480\times480$, and experiments reveal that the projection number $N_v=4$ can achieve the best balance between prediction efficiency and accuracy. We take the dimension of the latent variable $K_1=3$ and the channel size of intermediate features $K_2=32$ for a light computation overhead.

\subsection{Datasets \& Evaluation Metrics}

\textbf{Datasets.} Our experiments utilize three widely-used PCQA datasets, including SJTU-PCQA~\cite{yang2020predicting}, WPC~\cite{liu2022perceptual}, and WPC2.0~\cite{liu2021reduced}. We strictly follow the operations of previous works~\cite{fan2022no,zhang2022mm} and employ the k-fold cross-validation strategy to evaluate our method and other existing competitors. Specifically, we employ 9-fold, 5-fold, and 4-fold cross-validation for SJTU-PCQA, WPC, and WPC2.0, respectively, resulting in an approximate 8:2 split between training and testing sets. The final reported performance on a dataset is the average results across all folds. It is worth noting that all compared counterparts comply with the same dataset split to avoid any unfair comparisons.

\noindent\textbf{Evaluation Metrics.} We leverage four widely adopted evaluation metrics to quantify the level of agreement between predicted quality scores and ground truth MOSs. These metrics encompass the Spearman Rank Correlation Coefficient (SRCC), Pearson Linear Correlation Coefficient (PLCC), Kendall's Rank Correlation Coefficient (KRCC), and Root Mean Squared Error (RMSE). Since there can be value range misalignment between predicted results and ground truth MOSs, we follow the previous method~\cite{zhang2022mm} and introduce a common four-parameter logistic function to align their range. By employing the aforementioned four metrics, we can obtain convincing benchmarking results.  

\subsection{Comparisons with State-of-the-art Methods}
We compare our method with 15 representative methods, including 8 FR-PPCQA methods (\ie, MSE-p2po~\cite{mekuria2016evaluation}, HD-p2po~\cite{mekuria2016evaluation}, MSE-p2pl~\cite{tian2017geometric}, HD-p2pl~\cite{tian2017geometric}, PSNR-yuv~\cite{torlig2018novel}, PCQM~\cite{meynet2020pcqm}, GraphSIM~\cite{yang2020inferring}, and PointSSIM~\cite{alexiou2020towards}) and 7 NR-PCQA methods (\ie, BRISQUE~\cite{mittal2012no}, NIQE~\cite{mittal2012making}, IL-NIQE~\cite{zhang2015feature}, ResSCNN~\cite{liu2023point}, PQA-Net~\cite{liu2021pqa}, 3D-NSS~\cite{zhang2022nrqa}, and MM-PCQA~\cite{zhang2022mm}). The performance comparisons of different methods are illustrated in Table~\ref{tab1}. We can observe that: 1) Our method obtains the best performance among all NR-PCQA methods on all datasets, even surpassing the newly-proposed MM-PCQA~\cite{zhang2022mm} by a significant margin. For example, with the same experimental settings, our method outperforms the second-best MM-PCQA by 7.19\%  in terms of SRCC on the challenging WPC2.0 dataset. 2) Our method outperforms all FR-PCQA methods despite the additional reference information utilized by these methods. For instance, compared with the representative GraphSIM, our method shows 10.67\% SRCC improvement on the WPC2.0 dataset.  3) Our method exhibits robust performance (fewer performance fluctuations) on both the relatively easy SJTU-PCQA dataset and the challenging WPC2.0 dataset.

\subsection{Generalization Analyses}
Due to the data distribution variations and annotation biases in existing datasets, previous PCQA methods consistently suffer from poor cross-dataset generalization capabilities, which has been a persistent challenge for the practical application of existing PCQA algorithms. Our proposed method estimates point cloud quality by learning the conditional probability distribution of quality judgments in subjective tests rather than a solitary and potentially biased judgment, thus expected to have an excellent generalization capability. To compare the generalization performance of different methods, we select the relatively larger WPC (740 samples) as the training dataset and evaluate the models on the other two datasets, \ie, SJTU-PCQA (378 samples) and WPC2.0 (400 samples). Besides, since there exist reference overlaps between the WPC and WPC2.0 datasets, we remove the samples in WPC with overlapped references when testing on WPC2.0, to ensure a considerable cross-dataset generalization difficulty. 

The experimental results of cross-dataset generalization are listed in Table~\ref{tab2}, from which we can learn that the proposed method obtains the best generalization performance and exceeds other counterparts by a noticeable margin. In concrete terms, our method surpasses the existing best method MM-PCQA~\cite{zhang2022mm} by around 14.4\% and 6.9\% in terms of SRCC on the SJTU-PCQA and WPC2.0 datasets, respectively. Moreover, the compelling performance of our method even exceeds all existing approaches directly trained on the SJTU-PCQA and WPC2.0 datasets in Table~\ref{tab1}. Hence, this work offers a promising approach to addressing the enduring PCQA challenge on algorithm generalizability.

\subsection{Ablation Studies}
To study the effectiveness of our designs , we tune our model on the WPC2.0 dataset and analyze the performance changes.

\begin{table}[t]\small
\renewcommand{\tabcolsep}{2mm}
\centering
\begin{tabular}{l|cccc}
\toprule
Method & SRCC$\uparrow$       & PLCC$\uparrow$      & KRCC$\uparrow$    & RMSE$\downarrow$ \\
\midrule\midrule
MM-PCQA &0.8023 &0.8024 &0.6202  &0.1343 \\
\textbf{Ours}&\textbf{0.8742}&\textbf{0.8880}&\textbf{0.6922}&\textbf{0.0979} \\ \midrule
w/o Stochastic &0.7756 &0.7871 &0.5892 & 0.1341 \\
w/o KL Annealing & \ding{55} & \ding{55} & \ding{55} & \ding{55} \\
w/o $\mathcal{L}_{GSNN}$  &0.8478 &0.8643 &0.6531 &0.1092 \\
only $\mathcal{L}_{GSNN}$ &0.7229 &0.7583 &0.5356 &0.1410 \\
w/o Depth &0.8476 &0.8589 &0.6603 &0.1083 \\
Fixed Viewpoint & \ding{55} & \ding{55} & \ding{55} & \ding{55}\\
Early Average &0.8606 &0.8735 &0.6743 &0.1028\\
\bottomrule
\end{tabular}
\caption{Ablations of our method on WPC2.0. “\ding{55}” indicates an untrainable model due to gradient explosion.}  \label{tab3}
\end{table}

\noindent\textbf{Stochastic versus Deterministic.} 
To highlight the superiority of our probabilistic architecture over previous deterministic mapping-based approaches, we build a deterministic model that solely comprises the QRG and takes point cloud projections without stochastic features as input. As shown in Table~\ref{tab3}, the performance of the deterministic model, denoted as “w/o Stochastic,” exhibits noticeably inferior performance compared to our advanced probabilistic architecture. We attribute to the cause that traditional deterministic methods overlook the uncertain factors in subjective tests and are susceptible to biased ground truth annotations.

\begin{table}[t]\small
\renewcommand{\tabcolsep}{1.8mm}
\centering
\begin{tabular}{l|cccc}
\toprule
Projection Number & SRCC$\uparrow$       & PLCC$\uparrow$      & KRCC$\uparrow$    & RMSE$\downarrow$ \\
\midrule\midrule
2 & 0.8650 & 0.8837 & 0.6754 & 0.1002\\
\textbf{4} &\textbf{0.8742}&\textbf{0.8880}&\textbf{0.6922}&\textbf{0.0979} \\
6 &0.8498  &0.8696  &0.6630  &0.1039\\
\bottomrule
\end{tabular}
\caption{Ablation studies on the projection number.} 
\label{tab4}
\end{table}

\noindent\textbf{KL Annealing.}
Training a CVAE capable of generating diverse outputs necessitates training samples with multiple annotation versions. We adopt the KL annealing strategy to overcome the absence of diverse annotations from existing datasets. To prove the validity of our scheme, we further conduct experiments without the KL annealing strategy and find this approach makes the model become untrainable, indicating the necessity of the KL annealing strategy.

\noindent\textbf{GSNN Loss.} To showcase the efficacy of our GSNN loss, we explore two additional variants. The first variant merely encompasses the $\mathcal{L}_{CVAE}^{\lambda}$ term in Equation~(\ref{eq4}), referred to as “w/o $\mathcal{L}_{GSNN}$”. The second variant exclusively incorporates the $\mathcal{L}_{GSNN}$ term, denoted as “only $\mathcal{L}_{GSNN}$”. Experiments in Table~\ref{tab3} show that both variants yield suboptimal performance. The first variant fails to mitigate the distribution mismatch during training and testing, while the second variant invalids the learning of the posterior distribution.

\noindent\textbf{Depth Information.} 
Previous projection-based NR-PCQA methods~\cite{liu2021pqa,fan2022no,zhang2022mm} only utilize RGB colors in the point cloud projections, which cannot capture the spatial geometry distortions. To verify the benefits of additional depth information in our projections, we conduct a comparative experiment denoted as “w/o depth,” where the input projections only contain RGB colors. As shown in Table~\ref{tab3}, the absence of depth information indeed degrades the performance of models.

\noindent\textbf{Random Viewpoint Selection.} 
To simulate the interactive viewing in subjective tests, we introduce random viewpoint selection when computing point cloud projections. To investigate the role of our random viewpoint selection strategy, we perform ablation experiments with the fixed viewpoint strategy adopted by the previous work~\cite{zhang2022mm}. As demonstrated in Table~\ref{tab3} “Fixed Viewpoint”, using fixed viewpoints leads to an untrainable model, revealing the importance of our strategy in introducing stochastic priors and stabilizing the training process.

\noindent\textbf{Late Average versus Early Average.} During testing, our approach, marked as “Late Average,” samples the prior statistic multiple times to obtain stochastic features, each leading to a quality rating. We average all ratings of a sample to calculate the final quality prediction similar to the computation of MOSs. An alternative “Early Average” manner involves averaging the stochastic features first and then computing the final quality prediction on the averaged stochastic feature. As demonstrated in Table~\ref{tab3}, our “Late Average” scheme reflecting practical subjective tests can obtain better performance.

\begin{table}[t]\small
\renewcommand{\tabcolsep}{2.7mm}
\centering
\begin{tabular}{l|cccc}
\toprule
Value of $\alpha$ & SRCC$\uparrow$       & PLCC$\uparrow$      & KRCC$\uparrow$    & RMSE$\downarrow$ \\
\midrule\midrule
0.2 & 0.8686 & 0.8791 & 0.6832 & 0.0996\\
\textbf{0.4} &\textbf{0.8742}&\textbf{0.8880}&\textbf{0.6922}&\textbf{0.0979} \\
0.6 &0.8726  &0.8805  &0.6892  &0.0995\\
\bottomrule
\end{tabular}
\caption{Ablation experiments of different $\alpha$ settings.} 
\label{tab5}
\end{table}

\noindent\textbf{Projection Number.} To investigate the influence of the number of projections, as shown in Table~\ref{tab4}, we conduct experiments with varying projection numbers, showing that the projection number $N_v=4$ can achieve the best performance. Too few or many projections may cause insufficient or redundant information, eventually hindering the performance promotion. 

\noindent\textbf{Value of Hyperparameter $\alpha$.} Though both the adapted CAVE loss $\mathcal{L}_{CVAE}^{\lambda}$ and the GSNN loss $\mathcal{L}_{GSNN}$ are indispensable for our model to achieve its optimal performance, unbalanced ratio of these two terms in the objective function may discourage the performance. To find the most proper value of the weighting factor $\alpha$, we conduct ablation experiments on different values of $\alpha$. As shown in Table~\ref{tab5}, $\alpha=0.4$ can achieve the best balance between $\mathcal{L}_{CVAE}^{\lambda}$ and $\mathcal{L}_{GSNN}$.

\section{Conclusion}
This work presents the first probe of the stochasticity in dataset labeling for PCQA. To mimic the generation of ground truth MOSs, we propose a novel probabilistic architecture to model the conditional probability distribution of quality judgments in subjective tests. Specifically, our method utilizes a PCQA-tailored conditional variational autoencoder (CVAE) architecture to capture the uncertain variability in subjective tests and the deterministic distortions in point cloud projections, which is then integrated into diverse stochastic quality ratings for a point cloud. The stochastic ratings representing the labeling variants in subjective tests are finally averaged into an accurate quality prediction. Extensive experiments show that our method significantly outperforms previous approaches.

\section*{Acknowledgments}

This work was supported by The Major Key Project of PCL (PCL2024A02), Natural Science Foundation of China (62271013, 62031013), Guangdong Provincial Key Laboratory of Ultra High Definition Immersive Media Technology (2024B1212010006), Guangdong Province Pearl River Talent Program (2021QN020708), Shenzhen Science and Technology Program (JCYJ20240813160202004, JCYJ20230807120808017), Guangdong Basic and Applied Basic Research Foundation (2024A1515010155).

\bibliographystyle{named}
\bibliography{ijcai25}

\end{document}